\documentclass[letterpaper, 10 pt, conference]{ieeeconf}
\IEEEoverridecommandlockouts
\usepackage{comment}
\usepackage{amsmath} 
\usepackage{amssymb}  
\usepackage{graphicx}
\usepackage{amsfonts}
\usepackage{gensymb}
\usepackage{algorithmic}
\usepackage{textcomp}
\usepackage{xcolor}
\usepackage{graphicx}
\usepackage{dblfloatfix}    
\usepackage{pdfpages}

\usepackage{cite}
\makeatletter
\let\NAT@parse\undefined
\makeatother
\usepackage{hyperref}
\def\BibTeX{{\rm B\kern-.05em{\sc i\kern-.025em b}\kern-.08em
    T\kern-.1667em\lower.7ex\hbox{E}\kern-.125emX}}

\begin{document}

\title{A Comparison Between Joint Space and Task Space Mappings for Dynamic Teleoperation of an Anthropomorphic Robotic Arm in Reaction Tests \\
\author{Sunyu Wang, Kevin Murphy, Dillan Kenney, and Joao Ramos}
\thanks{The authors are with the Department of Mechanical Science and Engineering at the University of Illinois at Urbana-Champaign, USA. Corresponding author's contact: swang242@illinois.edu}
\thanks{The work is funded by the National Science Foundation via grant IIS-2024775.}
}
\maketitle

\begin{abstract}
Teleoperation---i.e., controlling a robot with human motion---proves promising in enabling a humanoid robot to move as dynamically as a human. But how to map human motion to a humanoid robot matters because a human and a humanoid robot rarely have identical topologies and dimensions. This work presents an experimental study that utilizes reaction tests to compare the proposed joint space mapping and the proposed task space mapping for dynamic teleoperation of an anthropomorphic robotic arm that possesses human-level dynamic motion capabilities. The experimental results suggest that the robot achieved similar and, in some cases, human-level dynamic performances with both mappings for the six participating human subjects. All subjects became proficient at teleoperating the robot with both mappings after practice, despite that the subjects and the robot differed in size and link length ratio and that the teleoperation required the subjects to move unintuitively. Yet, most subjects developed their teleoperation proficiencies more quickly with the task space mapping than with the joint space mapping after similar amounts of practice. This study also indicates the potential values of a three-dimensional task space mapping, a teleoperation training simulator, and force feedback to the human pilot for intuitive and dynamic teleoperation of a humanoid robot's arms. 
\end{abstract}


\section{Introduction}
\begin{figure}[t]
\centering
    \includegraphics[width = 0.999\linewidth]{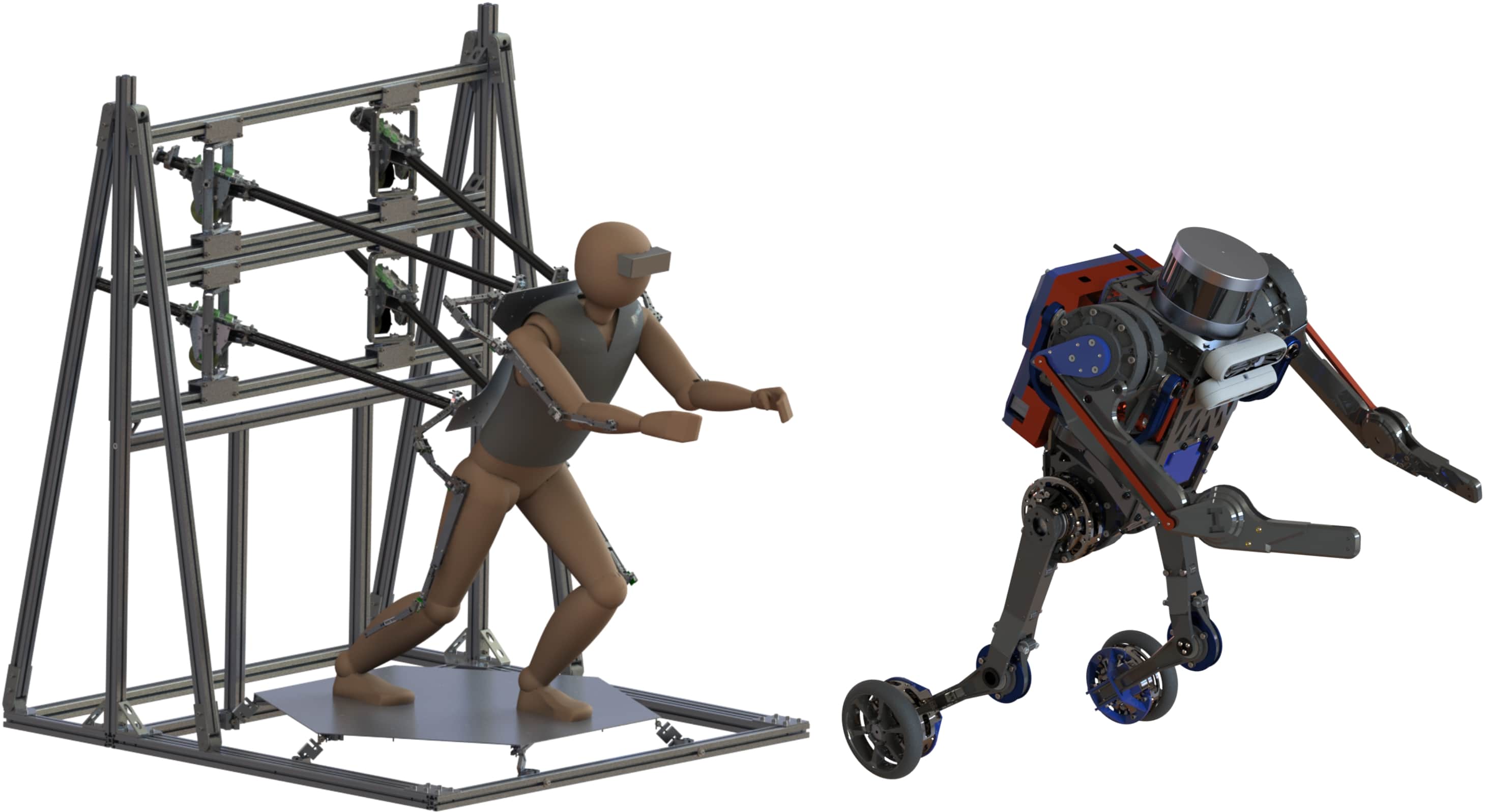}
    \caption{Design rendering of the envisioned human-humanoid robot system. Images are not in scale.}
    \label{fig:Human_and_SATYRR}
\end{figure}

Humans are amazing control systems and can outperform the state-of-the-art humanoid robots in dynamic motions. Hence, we aim to use teleoperation---i.e., controlling a robot with human motion---to grant human-level dynamic performance to humanoid robots. One application of humanoid robot teleoperation is disaster response, which inherently requires dynamic arm motion. Therefore, this work will focus on dynamic teleoperation of the robotic arm in our envisioned human-humanoid robot system shown in Fig. \ref{fig:Human_and_SATYRR}. 

There exist two most obvious ways of mapping human arm motion to a robot for teleoperation: 1) Joint space mapping, which synchronizes the human arm's joint positions with those of the robot. 2) Task space mapping, which synchronizes the human arm's Cartesian end-effector position with that of the robot via kinematic scaling. 

Existing works have demonstrated the potentials of teleoperation. In \cite{NAO} and \cite{One_Arm}, the respective human pilots can teleoperate the humanoid robot Nao to perform whole-body motions and a robotic arm on a wheeled base. Both systems use an inertial motion capture device and task space mapping. In \cite{Italian_Motion_Retargeting}, with the same type of motion capture system as in \cite{NAO} and \cite{One_Arm}, multiple robots of various topologies can follow the human pilot's complex motions, since the motion retargeting technique used is orientation-based. In \cite{MECHA_Unilateral} and \cite{MECHA_Bilateral}, the human pilot can teleoperate a human-size humanoid robot to carry load and perform simple manipulations with a laser-based motion capture system or a mechanical-linkage-based motion capture system. Yet, none of these robotic systems can realize the dynamic physical tasks which are trivial to humans, such as running, throwing, and heavy punching. 

In contrast, in \cite{HERMES}, the task space teleoperated HERMES's arm can move considerably faster and break through a wall barrier with mechanical-linkage-based motion capture. Hence, we assume two keys to dynamic teleoperation of a robotic arm: 1) The motion capture frequency must be sufficiently high ($\geq$ 1 kHz) for relatively fast positional update and accurate velocity and force estimations, which few existing tetherless motion capture systems can achieve. 2) The teleoperated robot must be capable of moving dynamically by design.

Besides hardware, the choice of mapping may also play a role in dynamic teleoperation of a robotic arm. Existing studies comparing different mappings tend to be system-specific. One of them focuses on robotically steered needle, a device for robotic surgery, and concludes that task space mapping yields quicker and more accurate needle insertion than joint space mapping \cite{Needle}. Another work compares the time required for a teleoperated robotic hand to pick and place different objects using different mappings \cite{Columbia_Hand}. In \cite{Croatia} and \cite{Delay}, a joint space control law and a task space one are employed for teleoperating the manipulators in the respective studies, but no specific application scenario or dynamic performance comparison is highlighted. 

None of the aforementioned works has investigated the mapping's effect on dynamic teleoperation of a humanoid robot's arm or utilized a system that attains human-level dynamic performance. Therefore, this work aims to fill the void: The contribution of this work is an experimental comparison of human's and robot's performances between the proposed joint space and the proposed task space mappings for dynamic teleoperation of an anthropomorphic robotic arm designed for performing dynamic motions, as shown in Fig. \ref{fig:Human_and_SATYRR}. The human's and the robot's dynamic performances are evaluated based on their reaction times in reaching tasks that require agility. The experimental results suggest that the robot achieved similar dynamic performances with both mappings after the six participating human subjects practiced the teleoperation. \emph{In some cases, the robot even performed the tasks as fast as the subjects performed the same tasks by themselves.} These results were obtained even though the subjects and the robot have different sizes and link length ratios, and the teleoperation required the subjects to move unintuitively. Yet, most subjects developed their teleoperation proficiencies faster with the task space mapping than with the joint space mapping after similar amounts of practice. 

\section{Methods}
\subsection{Motion Capture Linkage}
To capture human arm motion at high rates, a wearable motion capture linkage has been built. The linkage has nine degrees of freedom (DoFs) and length adjustment features, so its shoulder and elbow flexion/extension axes can be closely aligned with those of the human. Upon use, the human straps the linkage to the upper arm, so every human forearm configuration maps to a unique set of linkage joint positions. Seven of the nine linkage joint positions are sensed by magnetic encoders. The two DoFs at the connecting point to the human's forearm are not sensed as their purpose is to allow unconstrained forearm movement even if the linkage's and the human's elbow joint axes are not perfectly aligned. 

\subsection{Robotic Arm}
\begin{figure}[t]
\centering
    \includegraphics[width = 0.999\linewidth]{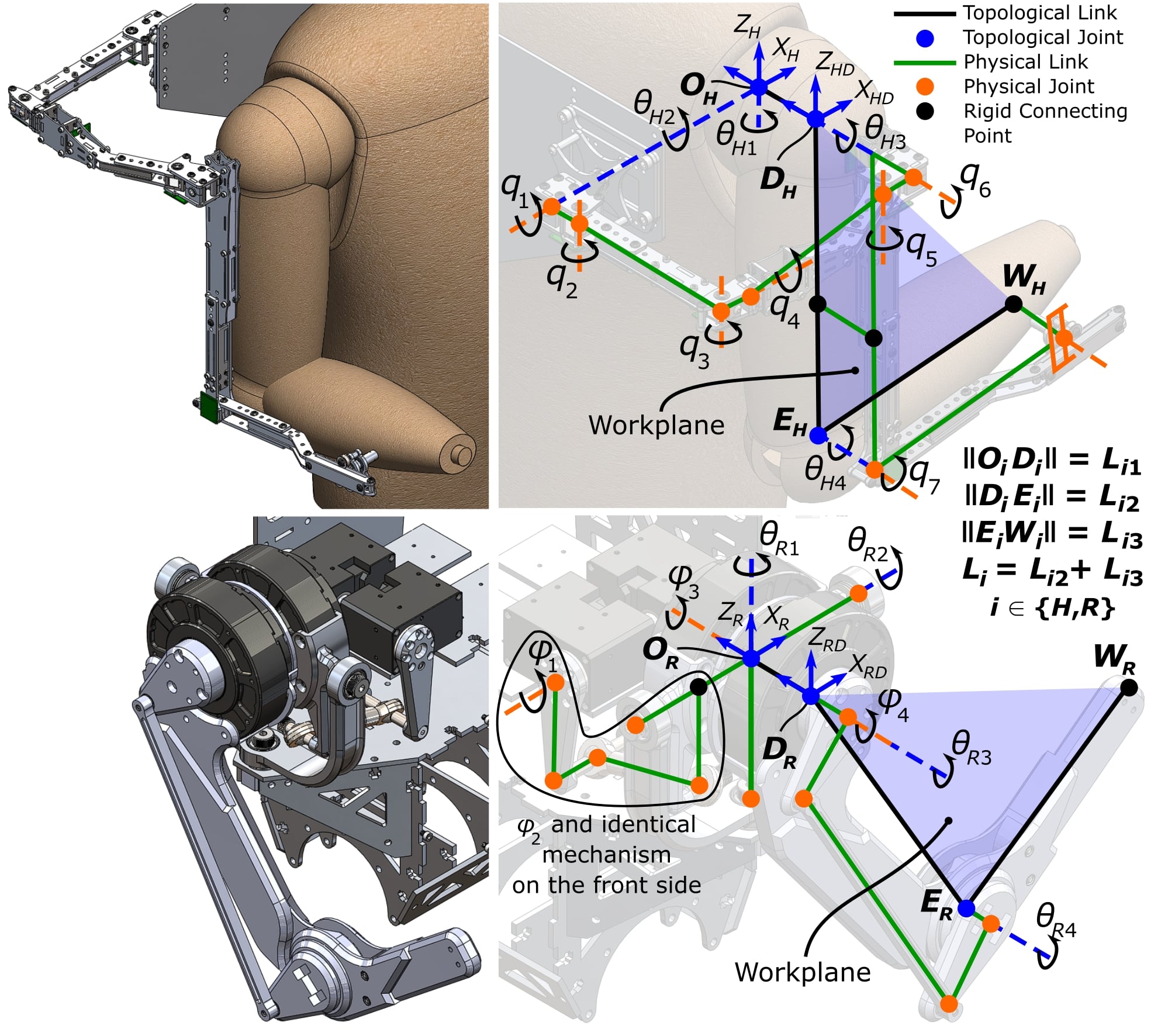}
    \caption{Mechanical designs and topological arm models of the human-linkage system and the robot. The human's and the robot's topological arm models share identical frame and joint position definitions, and only differ in link lengths. Subscript $H$ stands for human and subscript $R$ stands for robot. $O$ represents the shoulder joint and the origin of each system's inertial frame. $D$, $E$, and $W$ represent the deltoid, the elbow, and the wrist, respectively. Images are not in scale. }
    \label{fig:Hardware_Topological_Arm}
\end{figure}

The robotic arm used in this study is of four DoFs, approximately 50--60\% of a human adult's arm length, and has two DYNAMIXEL XM540 servo motors and two proprioceptive actuators \cite{Proprioceptive_Actuator}. As shown in Fig. \ref{fig:Hardware_Topological_Arm}, the two DYNAMIXEL XM540s and the shoulder gimbal form a differential mechanism actuating shoulder yaw $(\theta_{R1})$ and shoulder roll $(\theta_{R2})$. The proprioceptive actuator in the medial direction directly controls shoulder pitch $(\theta_{R3})$ and the other proprioceptive actuator controls elbow rotation $(\theta_{R4})$ through a parallelogram linkage. The selected actuators can achieve high actuation speed and their close placement to the shoulder reduces the robotic arm's inertia, ensuring its strong dynamic motion capabilities. Meanwhile, the robotic arm is of a high degree of anthropomorphism since it approximates human arm's joint topology. 

\subsection{Topological Arm Model}
If the wrist and the hand are ignored, several existing works model human arm as a 4-DoF chain composed of a spherical joint, a revolute joint, and two links in between \cite{Arm_IK_1}, \cite{Arm_IK_2}, \cite{Arm_IK_3}, \cite{Arm_IK_4}, \cite{Arm_IK_5}. In this work, the same model will be employed but with slight modification. The spherical joint will be separated into a universal joint that allows shoulder yaw $(\theta_{H1})$ and shoulder roll $(\theta_{H2})$, and a revolute joint that allows shoulder pitch $(\theta_{H3})$. Between these two joints, a third link $\left( \text{link } O_{H}D_{H} \text{ in Fig. \ref{fig:Hardware_Topological_Arm}} \right)$ is added to improve the model's fidelity. For the human arm, this added link accounts for the distance between the shoulder roll axis, which is roughly located in the lateral part of the scapula, and the connecting point between the deltoid and the motion capture linkage. For the robotic arm, this added link accounts for the thickness of the two proprioceptive actuators on the shoulder. 

To build the human's and the robot's topological arm models, first define the coordinate frame located at the each system's shoulder joint ($O_i \text{ in Fig. \ref{fig:Hardware_Topological_Arm}}$) as their respective inertial frames. Then, define the topological arm's rotational sequence as $\theta_{i1} \rightarrow \theta_{i2} \rightarrow \theta_{i3} \rightarrow \theta_{i4}, i \in \{H,R\}$. Since the motion capture linkage is strapped to the human's upper arm, $\theta_{H1}$ and $\theta_{H2}$ are computed via the relation: 
\begin{align} \label{eq:thH_from_q}
    R_z \left( \theta_{H1} \right) R_{x} \left( -\theta_{H2} \right) = R_{HOD} \left( q_1,q_2,q_3,q_4,q_5 \right),
\end{align}
where $R_z$ and $R_x$ are the standard rotation matrices about $z$- and $x$-axes, respectively, $R_{HOD}$ is $D_H$'s orientation in the human's inertial frame, and $q_k$'s are the linkage's encoder readings, $k \in \left[1, 5 \right]$. Since the linkage's shoulder pitch and elbow axes are closely aligned with those of the human arm, $q_6$ and $q_7$ are directly used as $\theta_{H3}$ and $\theta_{H4}$, respectively.

Because of the robotic arm's high degree of anthropomorphism, the proposed topological arm model is already equivalent to the robotic arm's topology.  

\subsection{Formulation and Implementation of Teleoperation}
Since this work focuses on teleoperation mappings, the robot controller will be kept on the lowest actuator joint level. Specifically, the human sends the robot topological joint position commands, which the robot translates to actuator joint position commands and executes a reactive position-velocity controller in the actuator joint space: 
\begin{align}
    \phi_{cmd} &= IK_{\phi\theta} \left( \theta_{cmd} \right), \label{eq:phicmd_from_thcmd} \\
    \dot{\phi}_{cmd} &= IK_{\phi\theta} \left( \dot{\theta}_{cmd} \right), \label{eq:phicmddot_from_thcmddot} \\
    \tau_{\phi} &= K_p \left( \phi_{cmd} - \phi \right) + K_d \left( \dot{\phi}_{cmd} - \dot{\phi} \right) \label{eq:robot_control},
\end{align}
where $\theta_{cmd} = \begin{bmatrix} \theta_{cmd1} & \theta_{cmd2} & \theta_{cmd3} & \theta_{cmd4}\end{bmatrix}^{\intercal}$ is the topological joint position commands the human sends to the robot, $\phi_{cmd} = \begin{bmatrix} \phi_{cmd1} & \phi_{cmd2} & \phi_{cmd3} & \phi_{cmd4}\end{bmatrix}^{\intercal}$ is the robot actuator joint position commands translated from $\theta_{cmd}$, $IK_{\phi\theta}(\cdot)$ is the corresponding inverse kinematics mapping, $\phi$ is the actual robot actuator joint positions, and $\tau_{\phi}$ is the torques exerted by the robot's actuators. 

With the robot controller and the topological arm model defined, joint space mapping is formulated as:
\begin{align}
    \theta_{cmd} = \theta_{H} \label{eq:joint_tele},
\end{align}
where $\theta_{H} = \begin{bmatrix} \theta_{H1} & \theta_{H2} & \theta_{H3} & \theta_{H4} \end{bmatrix}^{\intercal}$. For guarantee of feasibility of the command trajectory, task space mapping is formulated in a hybrid manner. First, define the plane formed by the topological upper arm and forearm as the workplane with the origin at the deltoid. Then, teleoperate $\theta_{R1}$ and $\theta_{R2}$ in joint space, whereas teleoperate $\theta_{R3}$ and $\theta_{R3}$ in the workplane's task space. Specifically, first scale the human's end-effector position in the human's workplane to that in the robot's workplane. Then, compute the latter half of $\theta_{cmd}$ with the inverse kinematics in the robot's workplane, where the robot is simply a planar 2-DoF manipulator: 
\begin{align} 
    \theta_{cmd1,2} &= \theta_{H1,2}, \label{eq:task_tele_1} \\
    \theta_{cmd3,4} &= IK_{\theta D} \left( \frac{L_R}{L_H}p_{HDW} \right) \label{eq:task_tele_2},
\end{align}
where $\theta_{H1,2} = \begin{bmatrix} \theta_{H1} & \theta_{H2} \end{bmatrix}^{\intercal}$, $\theta_{cmd1,2} = \begin{bmatrix} \theta_{cmd1} & \theta_{cmd2} \end{bmatrix}^{\intercal}$, $\theta_{cmd3,4} = \begin{bmatrix} \theta_{cmd3} & \theta_{cmd4} \end{bmatrix}^{\intercal}$, $p_{HDW}$ is the human's end-effector position in the human's workplane, $L_R$ and $L_H$ are the robot's and the human's respective arm lengths, and $IK_{\theta D}(\cdot)$ is the inverse kinematics mapping of the robot's upper arm and forearm in the robot's workplane. For both mappings, $\dot{\theta}_{cmd}$ is obtained by passing $\theta_{cmd}$ to a first-order velocity approximator with a cutoff frequency of 3 Hz.

This hybrid task space mapping is in fact an approximation because the element-wise ratio between the human's and the robot's end-effector positions in their respective inertial frames is not constant due to the presence of link $O_i D_i$, whose length is $L_{i1}$. Yet, if $L_{i1} = 0$, this inaccuracy would vanish. Hence, we assume that $L_{i1}$ is small compared with arm length $L_{i}$, so link $O_i D_i,i\in\{H,R\}$, will not significantly distort the precision of task space mapping. 

Two computers implement the proposed mappings at 1 kHz. The central computer, an NI cRIO-9082, first extracts the motion capture linkage's encoder readings, and computes $\theta_H$ and the human's end-effector position with the linkage's forward kinematics. Then, it executes (\ref{eq:joint_tele}) or (\ref{eq:task_tele_1}) and (\ref{eq:task_tele_2}) based on the user's choice of mapping, and sends $\theta_{cmd}$ and $\dot{\theta}_{cmd}$ to the robot's microcontroller, a Nucleo-F446RE, via a wired Serial Peripheral Interface communication protocol. Finally, the robot's microcontroller executes (\ref{eq:phicmd_from_thcmd}), (\ref{eq:phicmddot_from_thcmddot}), (\ref{eq:robot_control}), controlling the robot. The human pilot barely notices the communication delay or the robot's mechanical delay during teleoperation.

\begin{figure*}[!t]
\centering
    \includegraphics[width = 0.999\linewidth]{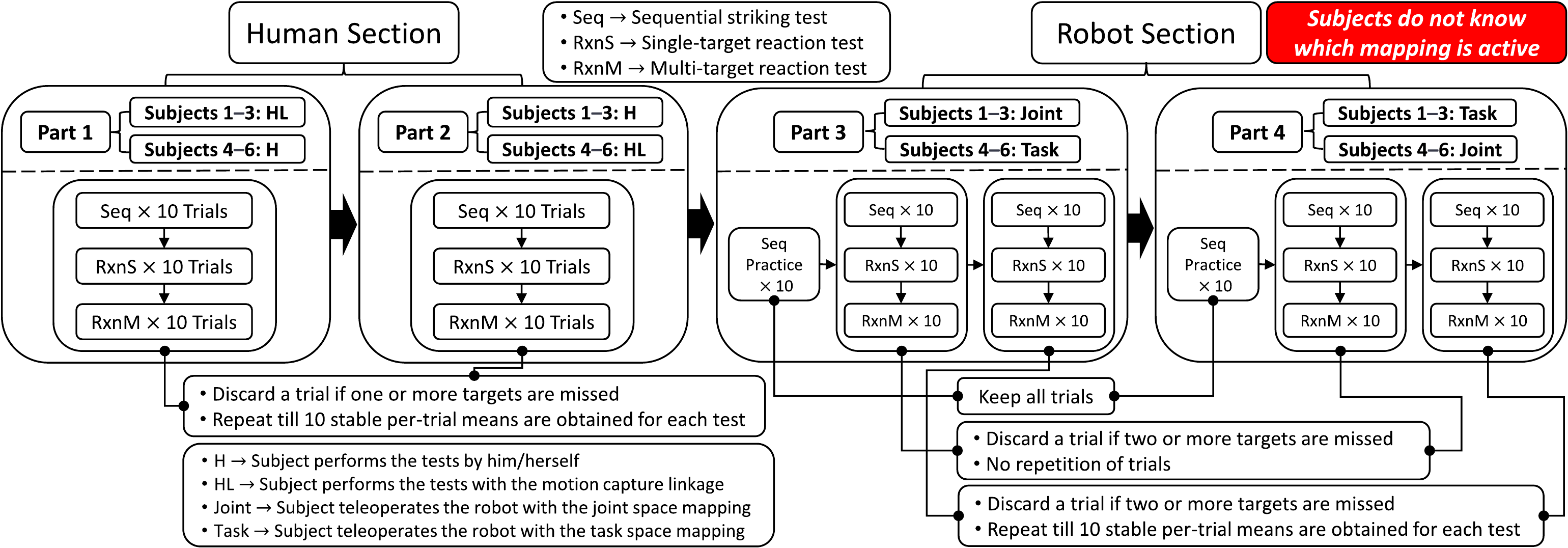}
    \caption{The experimental process for each subject. Three subjects perform each section in one order and the other three in the opposite order. The data logging method changes between the two experimental sections to obtain the subjects' stable reaction times and adaptation trends for each mapping.}
    \label{fig:Experimental_Process}
\end{figure*}

\subsection{Experimental Design}
With inspiration from the BATAK reaction test \cite{BATAK_Paper}, \cite{BATAK_Website}, an experiment involving six targets and six human subjects has been devised. Each target can be arbitrarily positioned in a vertical plane and consists of an LED light, a mechanical switch connected to a round button casing, and a proximity switch, which emits and receives infrared beam and changes its digital output signal if an object is detected within an adjustable distance to its infrared receiver. A target is triggered either when its button is pressed or when its proximity switch detects an object. This dual-switch design is to reduce the button-pressing action's influence on the experiment. 

\begin{figure}[t]
\centering
    \includegraphics[width = 0.999\linewidth]{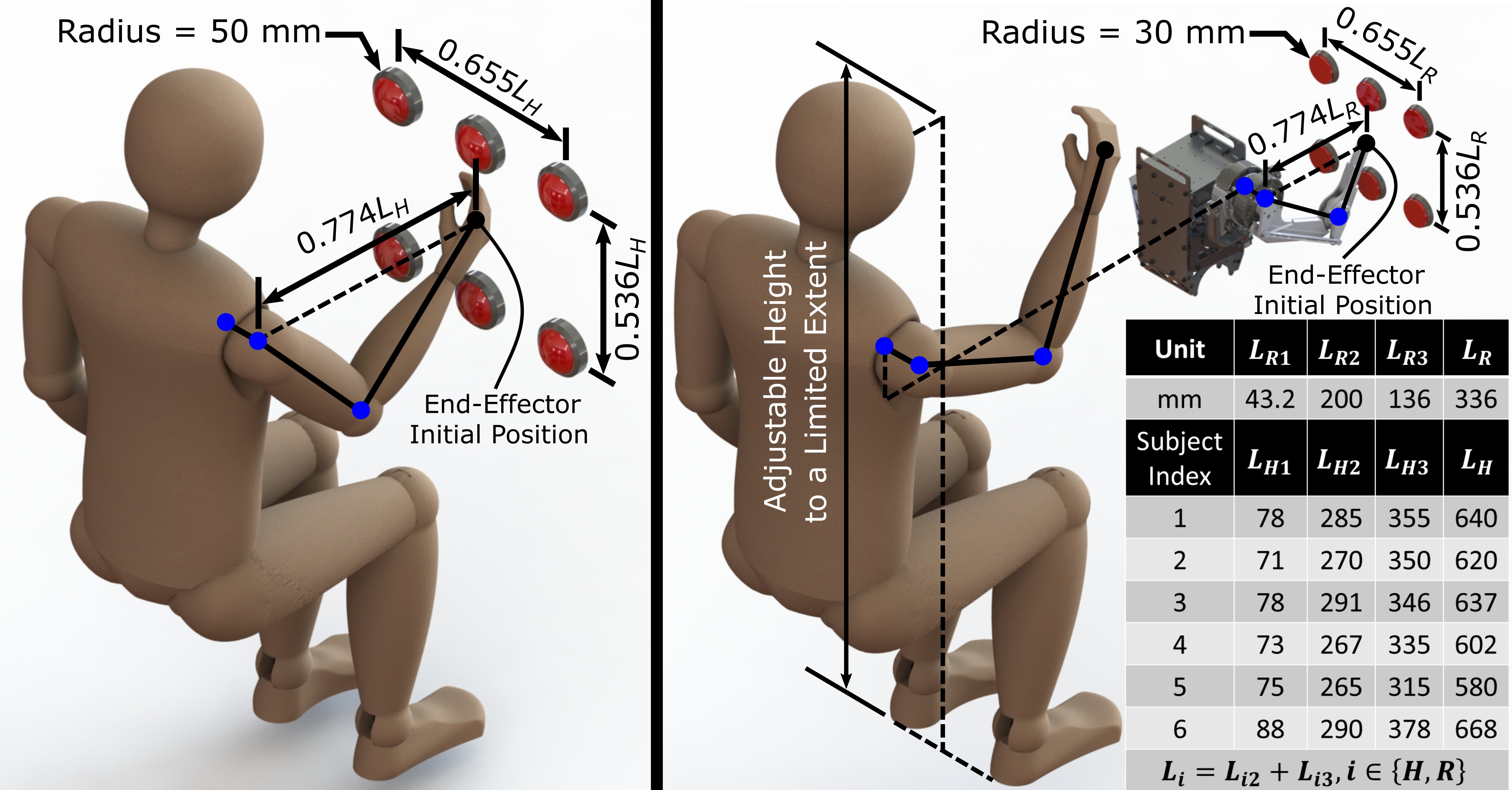}
    \caption{The experimental setup for each subject's human section (left) and robot section (right), which maintains constant positions of the targets relative to the subjects or the robot.}
    \label{fig:Experimental_Setup}
\end{figure}

One subject participates in the experiment at a time with the process shown in Fig. \ref{fig:Experimental_Process} and the setup in Fig. \ref{fig:Experimental_Setup}. The experiment centers around three tests: 1) Sequential striking test (Seq). 2) Single-target reaction test (RxnS). 3) Multi-target reaction test (RxnM). In a trial of sequential striking test, the subject will hit the top three targets from left to right and then the bottom three from right to left as fast as possible. The time between two adjacent hits is recorded, producing five reaction times per trial. In a trial of single-target reaction test, the top middle target lights up at a random time between 0.5--1 s after the previous hit, and the subject will hit the target as fast as possible after it lights up. A trial of multi-target reaction test is identical to a trial of single-target reaction test except that a random target lights up. For both reaction tests, the time from a target's light-up to its hit is recorded, a trial contains ten hits, and the subject returns to the end-effector initial position between hits. The central computer operates the targets at 8.3 kHz, the tests at 1 kHz, and logs data at 200 Hz. The supplementary video demonstrates the experiment with the physical setup. 

Before the experiment for each subject begins, the subject touches each target's center with the fist in the human section or with the robot's end-effector in the robot section. The targets' positions will be recorded and the proximity switches calibrated so that their triggerings are equivalent to punching the corresponding targets. Then, the subject receives the following instructions: 1) Minimize torso and wrist movements. 2) Do not predict a hit in the two reaction tests. 3) In the human section, punch the targets with the fist instead of pressing them with the fingers. 4) In the robot section, the priority order is safety $>$ accuracy $>$ speed, but pursue speed as soon as safety and accuracy are achieved comfortably. Lastly, the subject puts on an in-ear headphone playing white noise, and a section commences without intermission. Between the robot section's two parts, the subject is informed of the change of mapping, \emph{but does not know which mapping is active at any time during the section}. After the robot section ends, the subject is asked the following questions: 1) ``Which mapping do you prefer?" 2) ``What hindered you from moving faster?"

\begin{figure}[t]
\centering
    \includegraphics[width = 0.999\linewidth]{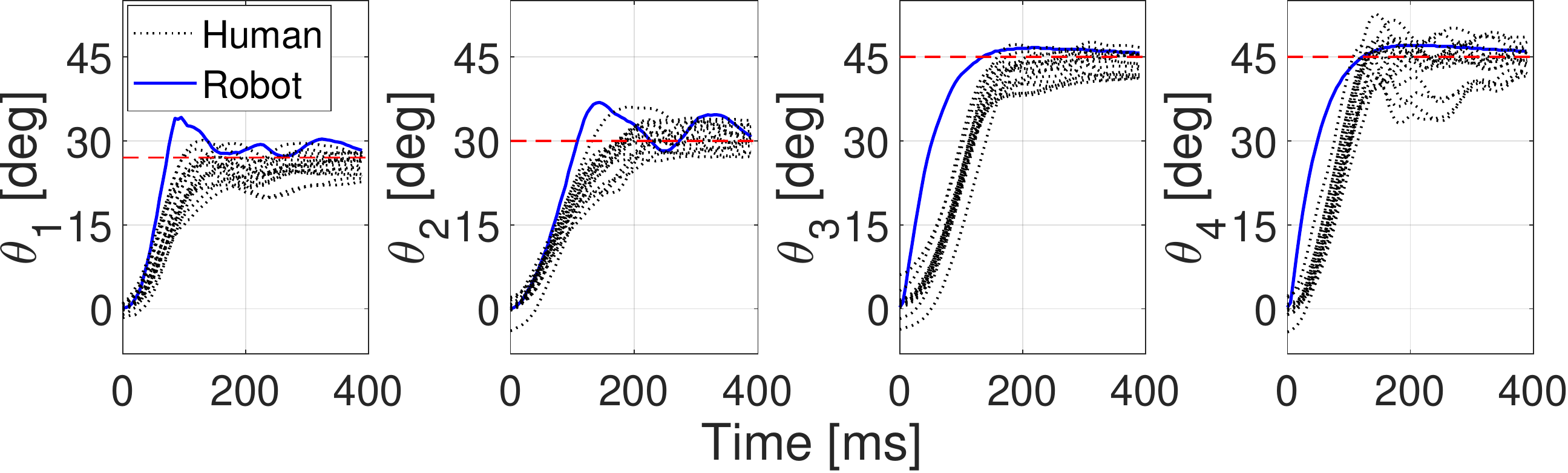}
    \caption{The robot controller's tuning results in each of the robot's four DoFs benchmarked by 16 step responses of the human subject who was the fastest at performing multi-target reaction test while wearing the motion capture linkage in the human section.}
    \label{fig:Step}
\end{figure}

A human section typically requires 1.5--2.5 hours to complete and a robot section 3--6 hours, as it involves more trials than a human section for the subject to practice teleoperating the robot with each mapping. Each subject performs the two sections on different days. After the last human section and before the first robot section of all subjects, the robot controller is tuned until the robot achieves a similar or faster step response in each of its four DoFs than the fastest human subject in multi-target reaction test with the motion capture linkage. The tuned robot controller remains unchanged in all robot sections, and the tuning results are shown in Fig. \ref{fig:Step}. 

\begin{figure*}[t]
\centering
    \includegraphics[width = 0.999\linewidth]{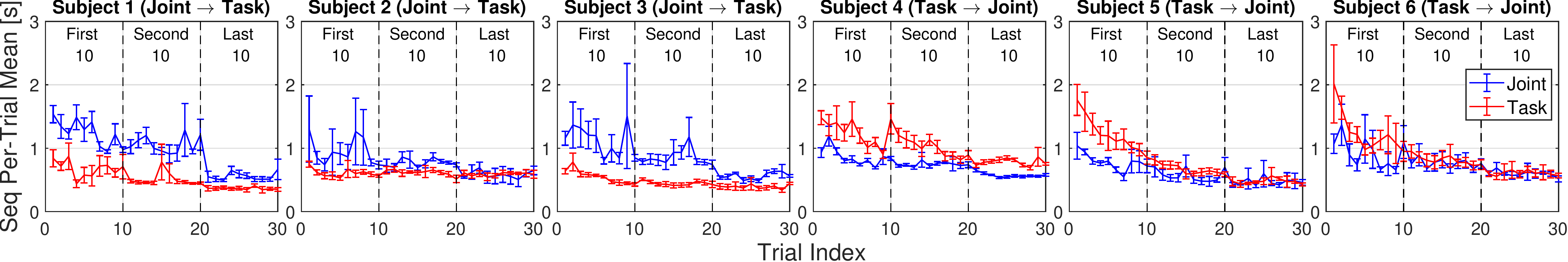}
    \caption{All subjects' per-trial mean reaction times with standard deviations in the first, the second, and the last ten trials of sequential striking test for each mapping.}
    \label{fig:Seq_Per_Trial}
\end{figure*}

\section{Results \& Discussion}
\begin{figure}[t]
\centering
    \includegraphics[width = 0.999\linewidth]{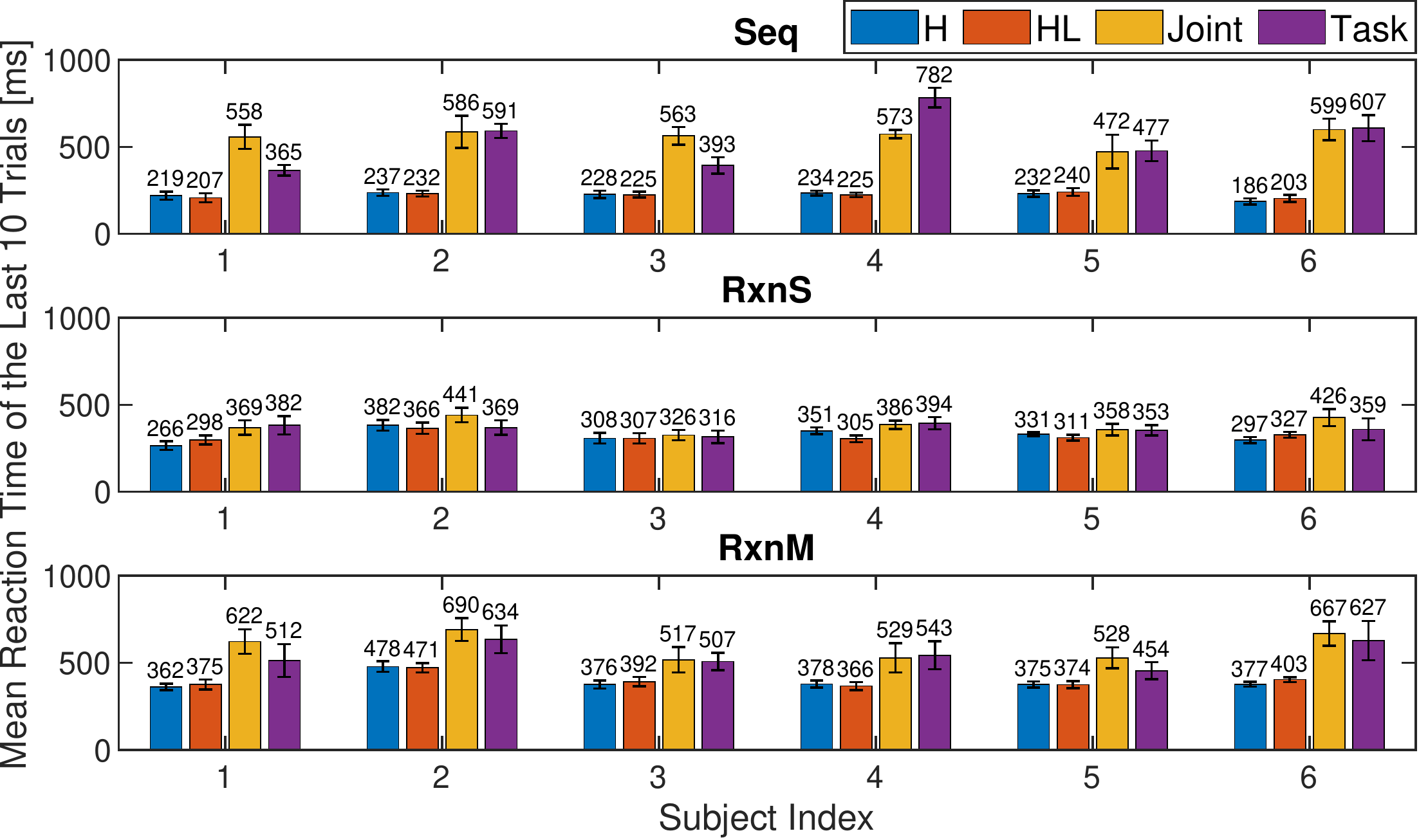}
    \caption{All subjects' stable mean reaction times in each testing combination.}
    \label{fig:Mean}
\end{figure}

\subsection{Stable Mean Reaction Times \& Adaptation Trends}
Fig. \ref{fig:Mean} shows the most important result of this study---all subjects' mean reaction times of the last ten trials in each testing combination, which represent the subjects' stable performances after adaptation in the experiment's time frame. Adaptation here means that, for one testing combination, a subject's per-trial mean reaction times and standard deviations decrease and converge as the number of trials increases. As shown in Fig. \ref{fig:Mean}, the stable mean reaction times for the two mappings vary by subjects in sequential striking test and are close in the two reaction tests. \emph{For single-target reaction test, the stable mean reaction times in all subjects' robot sections almost match those in their respective human sections.} The larger variations in sequential striking test could be explained by the fact that sequential striking test always occurs before the two reaction tests. Since a subject's extent of adaptation is measured by the trend of ten consecutive per-trial means, which does not account for the rate of adaptation, a subject could produce ten stable per-trial means but is still not fully adapted to the active mapping. 

Moreover, all subjects answered to the second post-robot-section question that, since the teleoperation system had no safety algorithm and the robot's behavior entirely depended on the subject, the concern of damaging the robot and the targets was the primary factor hindering the subjects from moving faster, especially during the first test for each mapping. The second most mentioned factor---by three subjects---was perception difficulty. These subjects could not comfortably see the bottom left and bottom middle targets because the extent of seat height adjustment was limited and the robot's torso partially blocked these targets. Two of these three subjects also could not accurately perceive if the robot's end-effector touched a target when they first started the teleoperation, and commented that developing the sense of depth based on visual feedback alone was challenging. These uncontrolled variables might have affected the subjects' adaptations and performances. 

Nevertheless, the similar stable mean reaction times for the two mappings in single-target and multi-target reaction tests suggest that all subjects were eventually fully adapted, and the choice of mapping did not significantly impact the robot's dynamic performances after the subjects' adaptations. 

However, subjects with different order of exposure to the two mappings adapted differently. Fig. \ref{fig:Seq_Per_Trial} shows each subject's per-trial mean reaction times with standard deviations in the first, the second, and the last ten trials of sequential striking test for each mapping. All subjects adapted to the mapping they began with. But after the mapping changed, subjects 2--3, who were exposed to the joint space mapping first and then the task space mapping, adapted to the second mapping more quickly and smoothly than subjects 4--6, who were exposed to the two mappings in the opposite order. The subjects' answers to the first post-robot-section question also reflect this phenomenon: Subjects 2--3 preferred the second mapping they used, and subjects 4--6 preferred the first mapping while commenting that teleoperating the robot with the second mapping they used was generally more difficult than with the first mapping. Subject 1's adaptation trend is an outlier because the subject experienced difficulty in concentrating due to fatigue after the mapping changed. 

\begin{figure*}[t]
\centering
    \includegraphics[width = 0.999\linewidth]{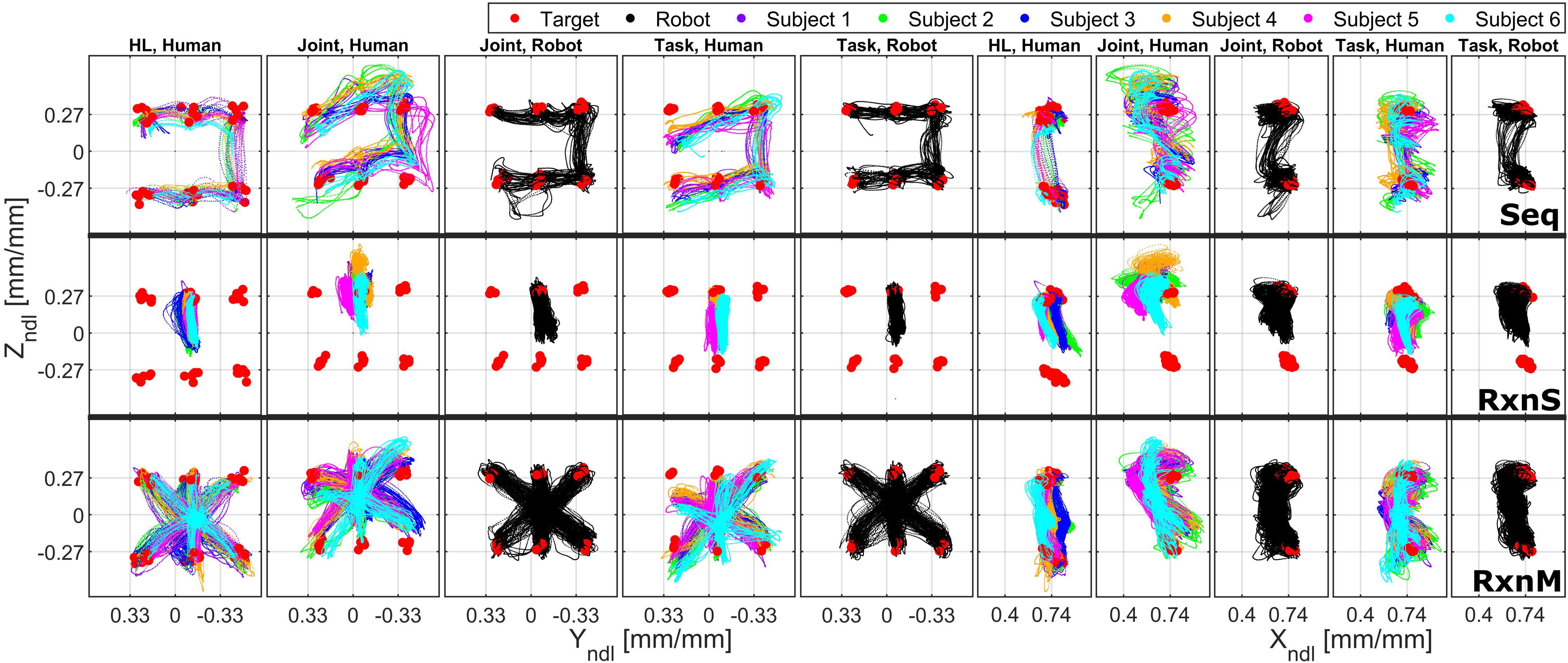}
    \caption{The subjects' and the robot's nondimensional end-effector trajectories in the last five trials of every testing combination where motion capture is available. The left 15 images show the frontal view and the right 15 the sagittal view. The top, middle, and bottom rows are for sequential striking, single-target reaction, and multi-target reaction tests, respectively. Subscript ``ndl" stands for ``nondimensional".}
    \label{fig:Nondimensional_Task_Space}
\end{figure*}

\subsection{End-Effector Trajectories in Nondimensional Task Space}
With actual length normalized by the subject's or the robot's arm length $L_{i}, i\in\{ H, R \}$, Fig. \ref{fig:Nondimensional_Task_Space} demonstrates the subjects' and the robot's nondimensional end-effector trajectories in the last five trials of every testing combination where motion capture is available. Two features stand out: 1) For the joint space mapping, the subjects' end-effector trajectories are less straight and higher than those of the robot. For the task space mapping, the subjects' and the robot's trajectories match more closely. 2) For both mappings, the subjects' and the robot's trajectories have certain orientational offsets. 

\begin{figure}[t]
\centering
    \includegraphics[width = 0.999\linewidth]{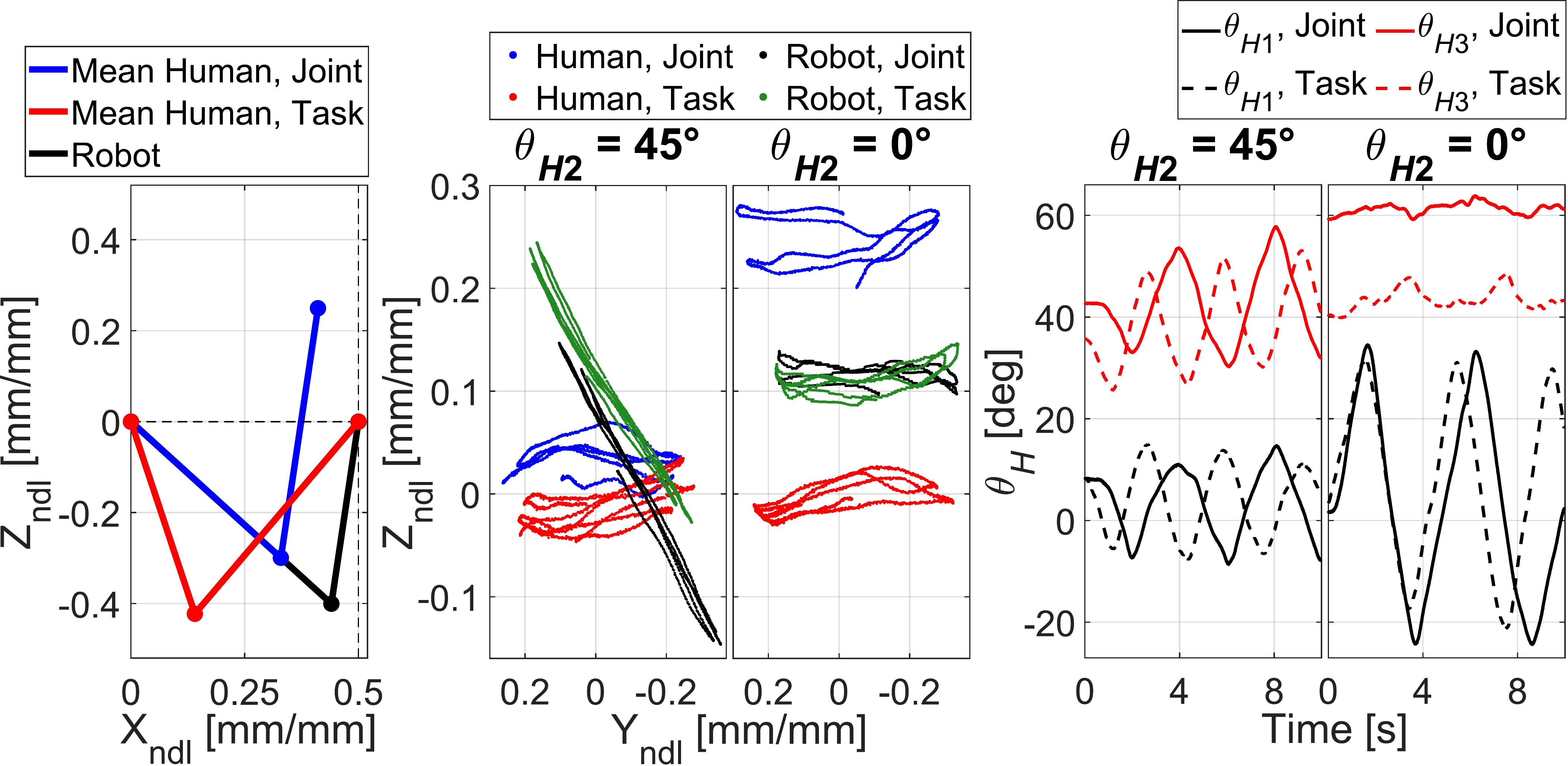}
    \caption{Explanations of the two features that stand out in Fig. \ref{fig:Nondimensional_Task_Space}. Images are not in scale.}
    \label{fig:Fig_9}
\end{figure}

The leftmost image in Fig. \ref{fig:Fig_9}, where the mean human arm is constructed with the six subjects' mean upper arm and forearm lengths, explains the height difference in the first feature. Since all subjects' forearms are longer than their upper arms whereas the robot's forearm is shorter than its upper arm, for the robot to reach a certain nondimensional height, all subjects need to reach higher with the joint space mapping than with the task space mapping. 

The second feature is explained by another experiment, where a subject moves the forearm horizontally in the frontal plane with $\theta_{H2} = 45\degree$ and then with $\theta_{H2} = 0\degree$ while teleoperating the robot with each mapping. As shown in the middle image in Fig. \ref{fig:Fig_9}, the orientational offset is pronounced when $\theta_{H2} = 45\degree$ but small when $\theta_{H2} = 0\degree$ for both mappings. Meanwhile, the rightmost image in Fig. \ref{fig:Fig_9} reveals that $\theta_{H1}$ and $\theta_{H3}$ change by similar magnitudes when $\theta_{H2} = 45\degree$, but $\theta_{H1}$ changes considerably more and $\theta_{H3}$ considerably less when $\theta_{H2} = 0\degree$. For the topological arm model's end-effector to move horizontally in the frontal plane, $\theta_{i1}$ must change. But if $\theta_{i1}$ and $\theta_{i3}$ change by similar magnitudes, the end-effector will move more drastically in the workplane---which is oriented diagonally when $\theta_{i2}$ is large---than the deltoid moves in the frontal plane, because $L_{i}$ is larger than $L_{i1}$. Hence, the combination of three factors causes the orientational offsets in Fig. \ref{fig:Nondimensional_Task_Space}: 1) The motion capture linkage's design, which makes the linkage capture $\theta_{H3}$ more sensitively when $\theta_{H2}$ is large. 2) The topological arm model, which relies on $\theta_{i1}, i\in\{H,R\}$, for horizontal end-effector movement in the frontal plane. 3) The two mappings, which always joint space teleoperate $\theta_{R1}$ and $\theta_{R2}$. 

Nevertheless, Fig. \ref{fig:Nondimensional_Task_Space} suggests that all subjects changed their behaviors in the robot section compared with the human section and adapted to both features despite their unintuitiveness. 

\subsection{Limitations of This Study}
This study's primary limitation is the relatively small sample size of six subjects due to the experiment's significant time consumption and strenuousness. Secondly, part of the experimental results might have been affected by uncontrolled variables, including the subjects': 1) Concern of damaging the teleoperation system due to its lack of safety algorithm. 2) Potentially different extents of adaptation in sequential striking test due to the experimental design's not accounting for the subjects' different adaptation rates. 3) Perception differences due to the experimental setup's limited extent of seat height adjustment. 4) Different concentration levels due to fatigue and loss of incentive, as the subjects' performances were not rewarded. 

Yet, these limitations do not degrade the value of this study. 

\section{Conclusion \& Future Work}
The contribution of this work is an experimental comparison of human's and robot's reaction times between the proposed joint space and the proposed task space mappings for dynamic teleoperation of an anthropomorphic robotic arm with human-level dynamic motion capabilities. The conclusion is that the robot achieved similar reaction times with both mappings for all six participating human subjects after the subjects adapted to each mapping. In addition, \emph{the robot performed single-target reaction test almost as fast as the subjects performed the same test by themselves.} These results were obtained even though: 1) The subjects and the robot differed in size and link length ratio. 2) The teleoperation required the subjects to move their arms differently than when they performed the same tests by themselves. However, most subjects adapted to the task space mapping faster than the joint space mapping after similar amounts of practice.

The conclusion and the limitations of this study indicate the following directions for future work: 1) A three-dimensional task space mapping that task space teleoperates all DoFs of the robot may eliminate the orientational offset between the human's and the robot's nondimensional end-effector trajectories and reveal more differences between the joint space and the task space mappings. 2) Training a human in a teleoperation simulator where the cost of damaging the robot or the robot's environment is negligible may accelerate the human's adaptation to the teleoperation compared with training the human with the physical robot. 3) Providing more than visual feedback to the human---such as providing force feedback---may assist the teleoperation. 

\section*{Acknowledgement}
The corresponding author would like to sincerely thank Professor Sangbae Kim for donating the central computer used in this study, Amartya Purushottam for the vital electronics support, and all subjects for their commitment and perseverance during the experiment. 

\bibliographystyle{IEEEtran}
\bibliography{ICRA_2021_References.bib}

\begin{thebibliography}{10}
\providecommand{\url}[1]{#1}
\csname url@samestyle\endcsname
\providecommand{\newblock}{\relax}
\providecommand{\bibinfo}[2]{#2}
\providecommand{\BIBentrySTDinterwordspacing}{\spaceskip=0pt\relax}
\providecommand{\BIBentryALTinterwordstretchfactor}{4}
\providecommand{\BIBentryALTinterwordspacing}{\spaceskip=\fontdimen2\font plus
\BIBentryALTinterwordstretchfactor\fontdimen3\font minus
  \fontdimen4\font\relax}
\providecommand{\BIBforeignlanguage}[2]{{%
\expandafter\ifx\csname l@#1\endcsname\relax
\typeout{** WARNING: IEEEtran.bst: No hyphenation pattern has been}%
\typeout{** loaded for the language `#1'. Using the pattern for}%
\typeout{** the default language instead.}%
\else
\language=\csname l@#1\endcsname
\fi
#2}}
\providecommand{\BIBdecl}{\relax}
\BIBdecl

\bibitem{NAO}
J.~{Koenemann}, F.~{Burget}, and M.~{Bennewitz}, ``Real-time imitation of human
  whole-body motions by humanoids,'' in \emph{2014 IEEE International
  Conference on Robotics and Automation (ICRA)}, 2014, pp. 2806--2812.

\bibitem{One_Arm}
M.~Arduengo, A.~Arduengo, A.~Colomé, J.~Lobo-Prat, and C.~Torras, ``A robot
  teleoperation framework for human motion transfer,'' 2019.

\bibitem{Italian_Motion_Retargeting}
K.~{Darvish}, Y.~{Tirupachuri}, G.~{Romualdi}, L.~{Rapetti}, D.~{Ferigo},
  F.~J.~A. {Chavez}, and D.~{Pucci}, ``Whole-body geometric retargeting for
  humanoid robots,'' in \emph{2019 IEEE-RAS 19th International Conference on
  Humanoid Robots (Humanoids)}, 2019, pp. 679--686.

\bibitem{MECHA_Unilateral}
Y.~{Ishiguro}, K.~{Kojima}, F.~{Sugai}, S.~{Nozawa}, Y.~{Kakiuchi}, K.~{Okada},
  and M.~{Inaba}, ``High speed whole body dynamic motion experiment with real
  time master-slave humanoid robot system,'' in \emph{2018 IEEE International
  Conference on Robotics and Automation (ICRA)}, 2018, pp. 5835--5841.

\bibitem{MECHA_Bilateral}
Y.~{Ishiguro}, T.~{Makabe}, Y.~{Nagamatsu}, Y.~{Kojio}, K.~{Kojima},
  F.~{Sugai}, Y.~{Kakiuchi}, K.~{Okada}, and M.~{Inaba}, ``Bilateral humanoid
  teleoperation system using whole-body exoskeleton cockpit tablis,''
  \emph{IEEE Robotics and Automation Letters}, vol.~5, no.~4, pp. 6419--6426,
  2020.

\bibitem{HERMES}
A.~{Wang}, J.~{Ramos}, J.~{Mayo}, W.~{Ubellacker}, J.~{Cheung}, and S.~{Kim},
  ``The hermes humanoid system: A platform for full-body teleoperation with
  balance feedback,'' in \emph{2015 IEEE-RAS 15th International Conference on
  Humanoid Robots (Humanoids)}, 2015, pp. 730--737.

\bibitem{Needle}
A.~{Majewicz} and A.~M. {Okamura}, ``Cartesian and joint space teleoperation
  for nonholonomic steerable needles,'' in \emph{2013 World Haptics Conference
  (WHC)}, 2013, pp. 395--400.

\bibitem{Columbia_Hand}
C.~{Meeker}, T.~{Rasmussen}, and M.~{Ciocarlie}, ``Intuitive hand teleoperation
  by novice operators using a continuous teleoperation subspace,'' in
  \emph{2018 IEEE International Conference on Robotics and Automation (ICRA)},
  2018, pp. 5821--5827.

\bibitem{Croatia}
\BIBentryALTinterwordspacing
G.~Vasiljevic, N.~Jagodin, and Z.~Kovacic, ``Kinect-based robot teleoperation
  by velocities control in the joint/cartesian frames,'' \emph{IFAC Proceedings
  Volumes}, vol.~45, no.~22, pp. 805 -- 810, 2012, 10th IFAC Symposium on Robot
  Control. [Online]. Available:
  \url{http://www.sciencedirect.com/science/article/pii/S1474667016337089}
\BIBentrySTDinterwordspacing

\bibitem{Delay}
P.~M. {Kebria}, A.~{Khosravi}, S.~{Nahavandi}, A.~{Homaifar}, and M.~{Saif},
  ``Experimental comparison study on joint and cartesian space control schemes
  for a teleoperation system under time-varying delay,'' in \emph{2019 IEEE
  International Conference on Industrial Technology (ICIT)}, 2019, pp.
  108--113.

\bibitem{Proprioceptive_Actuator}
P.~M. {Wensing}, A.~{Wang}, S.~{Seok}, D.~{Otten}, J.~{Lang}, and S.~{Kim},
  ``Proprioceptive actuator design in the mit cheetah: Impact mitigation and
  high-bandwidth physical interaction for dynamic legged robots,'' \emph{IEEE
  Transactions on Robotics}, vol.~33, no.~3, pp. 509--522, 2017.

\bibitem{Arm_IK_1}
A.~M. {Zanchettin}, P.~{Rocco}, L.~{Bascetta}, I.~{Symeonidis}, and
  S.~{Peldschus}, ``Kinematic motion analysis of the human arm during a
  manipulation task,'' in \emph{ISR 2010 (41st International Symposium on
  Robotics) and ROBOTIK 2010 (6th German Conference on Robotics)}, 2010, pp.
  1--6.

\bibitem{Arm_IK_2}
S.~{Parasuraman}, K.~C. {Yee}, and A.~{Oyong}, ``Human upper limb and arm
  kinematics for robot based rehabilitation,'' in \emph{2009 IEEE/ASME
  International Conference on Advanced Intelligent Mechatronics}, 2009, pp.
  845--850.

\bibitem{Arm_IK_3}
M.~Mihelj, ``Human arm kinematics for robot based rehabilitation,''
  \emph{Robotica}, vol.~24, no.~3, p. 377–383, 2006.

\bibitem{Arm_IK_4}
A.~Bertomeu-Motos, A.~Blanco, F.~Badesa, J.~Barios, L.~Zollo, and N.~Garcia,
  ``Human arm joints reconstruction algorithm in rehabilitation therapies
  assisted by end-effector robotic devices,'' \emph{Journal of NeuroEngineering
  and Rehabilitation}, vol.~15, 12 2018.

\bibitem{Arm_IK_5}
E.~{Papaleo}, L.~{Zollo}, S.~{Sterzi}, and E.~{Guglielmelli}, ``An inverse
  kinematics algorithm for upper-limb joint reconstruction during robot-aided
  motor therapy,'' in \emph{2012 4th IEEE RAS EMBS International Conference on
  Biomedical Robotics and Biomechatronics (BioRob)}, 2012, pp. 1983--1988.

\bibitem{BATAK_Paper}
D.~Gierczuk and Z.~Bujak, ``Reliability and accuracy of batak lite tests used
  for assessing coordination motor abilities in wrestlers,'' \emph{Polish
  Journal of Sport and Tourism}, vol.~21, 01 2014.

\bibitem{BATAK_Website}
\BIBentryALTinterwordspacing
R.~Wood, ``Batak light board reaction test,'' 2008, accessed 10/29/2020.
  [Online]. Available:
  \url{https://www.topendsports.com/testing/tests/reaction-batak.htm}
\BIBentrySTDinterwordspacing

\end{thebibliography}

\end{document}